\documentclass{article}
\usepackage{spconf,amsmath,amsthm,graphicx}
\usepackage{amsbsy}
\usepackage{url}
\usepackage{subfigure}

\newcommand{\mb}[1]{\mathbf{#1}}

\newcommand{\mc}[1]{\mathcal{#1}}

\usepackage{epstopdf}
\begin{document}
\title{Automatic Classification of Irregularly Sampled Time Series with Unequal Lengths: A Case Study on Estimated Glomerular Filtration Rate}

\name{Santosh Tirunagari, Simon Bull, Norman Poh\thanks{This work was supported by the Medical Research Council [grant number MR/M023281/1]. We would like to thank our collaborators at East Kent Hospital, UK for providing us with the dataset. The project details can be found at \scriptsize{http://www.modellingckd.org/} }}
\address{Department of Computer Science, University of Surrey, Guildford, Surrey GU2 7XH.\\ \{santosh.tirunagari, s.c.bull, n.poh\}@surrey.ac.uk}



\maketitle

\begin{abstract}
A patient's estimated glomerular filtration rate (eGFR) can provide important information about disease progression and kidney function. Traditionally, an eGFR time series is interpreted by a human expert labelling it as stable or unstable. While this approach works for individual patients, the time consuming nature of it precludes the quick evaluation of risk in large numbers of patients. However, automating this process poses significant challenges as eGFR measurements are usually recorded at irregular intervals and the series of measurements differs in length between patients. Here we present a two-tier system to automatically classify an eGFR trend. First, we model the time series using Gaussian process regression (GPR) to fill in `gaps' by resampling a fixed size vector of fifty time-dependent observations. Second, we classify the resampled eGFR time series using a K-NN/SVM classifier, and evaluate its performance via 5-fold cross validation. Using this approach we achieved an F-score of $0.90$, compared to $0.96$ for 5 human experts when scored amongst themselves.
\end{abstract}



\section{Introduction}

\begin{figure}[ht]
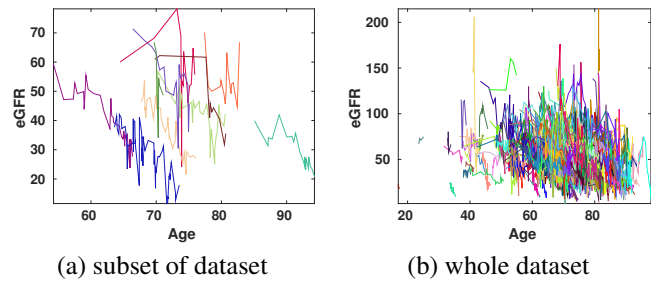

\centering
\begin{tabular}{cc}
\includegraphics[scale=0.22]{Pictures_B/problem.eps} & \includegraphics[scale=0.22]{Pictures_B/dataset.eps} \\
(a) subset of dataset & (b) whole dataset\\
\end{tabular}
\caption{A patient's eGFR is observed at irregular time intervals and over different age ranges. Each colour represents a single patient.}
\label{problem}
\end{figure}

Estimated glomerular filtration rate (eGFR) is a derived measurement that characterises the effective functioning of a kidney. It plays a central role in both the management of people with chronic diseases and epidemiology research involving longitudinal data with ten or more years of observations~\cite{poh2012calibrating}. Often eGFR time series exhibit irregularities such as missing values and unequal lengths. Missing values are an inevitable consequence of the difficulty of ensuring that patients return for regular follow up measurements~\cite{poh2014challenges}, while unequal lengths are a result of patients with differing ages and conditions receiving measurements with different frequency (Figure~\ref{problem}). A patient's eGFR is therefore observed at irregular time intervals, and will have greater or fewer observations depending on their age and the conditions they suffer from.

For a clinician, having an easily understandable summary of a patient's eGFR trend can be useful for determining the progression of diseases ranging from diabetes to chronic kidney disease. In many cases, simply distinguishing between stable (non-decreasing) and unstable (decreasing) trends can prove sufficient. Armed with this information a clinician can identify those patients who are most at risk of suffering a deterioration in their renal function. Presently, this trend differentiation is performed by a nephrologist manually analysing and labelling an eGFR time series. Despite manual labelling being time consuming and expensive, automating the process using standard supervised classifiers directly is not possible, as they require an equal number of input features and eGFR time series are of unequal length. Overcoming this requires either developing a framework for classifying the irregular and unequal eGFR time series, or using interpolation to make the time series amenable to standard classification methods. 

Prior work in time series analysis has strongly emphasised regularly sampled equal length series, resulting in fewer methods that exist specifically for analysing irregularly sampled or unequal time series data. If the time series is to be analysed directly, then approaches such as spectral analysis~\cite{schulz1997spectrum,stoica2011new} and kernel-based methods~\cite{rehfeld2011comparison} have been used to extract causal structure or statistics from data in fields such as astronomy~\cite{broersen2008spectral, thiebaut2005irregular}, palaeontology~\cite{rehfeld2011comparison} and economics~\cite{muller2000ema}. Despite this, the most common approach when working with irregular time series data is to transform it into regularly spaced data through some form of interpolation. For classification tasks, this has the added advantage of enabling the time series to be equalised to a given length by sampling from the interpolated function, thereby enabling standard classification algorithms to be used. 

In this study, classification was performed after equalising eGFR series lengths using Gaussian processes, which are particularly useful for predicting patient outcomes~\cite{windridge2014kernel}, and have been used previously when classifying irregular time series data~\cite{marlin2015GPR}. By enabling longitudinal data-interpolation and data-extrapolation, Gaussian processes can also directly assist with the problems of missing observations (irregular samples) and unequal series lengths respectively. Therefore, we employed Gaussian process regression (GPR) to equalise the lengths of the eGFR time series across all patients by producing the best linear unbiased predictions of interpolated and extrapolated observations. Following this equalisation, we used K-NN/SVM to classify an eGFR time series as having a stable or unstable trend.

The main objectives of this study are two-fold. First, to examine the feasibility of using machine learning based techniques, such as GPR, for transforming an unequal length irregularly sampled time series to an equal length, and then automatically classifying the equalised time series using a standard classifier. Second, to understand whether the trends classified by our approach correspond to those expected by clinicians. Our contributions can thus be summarised as follows: (i) Novel use of GPR+K-NN/SVM for classifying irregular eGFR time series with unequal lengths. 
Although GPR has been widely used, its uses in sampling irregular time series with unequal lengths are rarely highlighted or discussed. Our approach represents the first time that individual patient's eGFR time series have had their trend automatically classified and visualised, and serves to demonstrate the applicability of GPR to such problems. (ii) Improved understanding for clinicians. Through comparison with experts' trend classifications, we will demonstrate that our approach can serve as a tool for enhancing clinicians diagnostic capabilities. 

The organisation of the paper is as follows: In section~\ref{data}, we study the eGFR dataset. In section~\ref{method}, we describe our methodological framework. Experiments and results are discussed in section~\ref{exp}. Finally, in section~\ref{conclusion}, we draw conclusions and discuss their clinical relevance. 

\section{The `Hannah' Dataset}
\label{data}
\begin{figure}[ht]
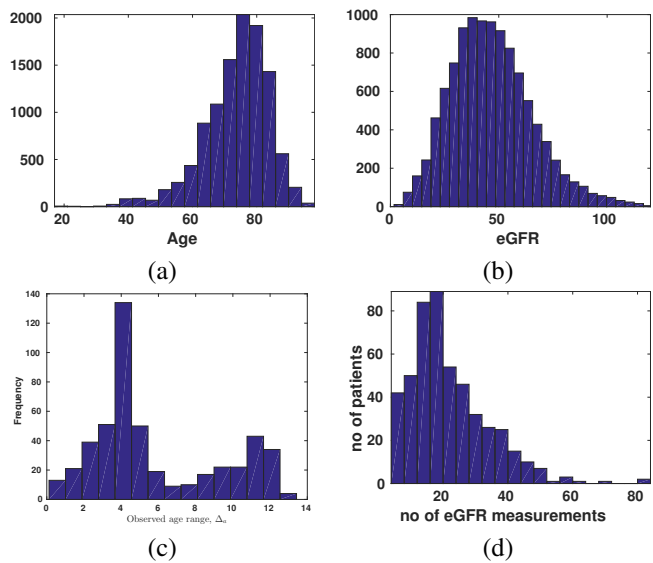

\centering
\begin{tabular}{cc}
\includegraphics[scale=0.22]{Pictures_B/age_histogram} & \includegraphics[scale=0.22]{Pictures_B/egfr_histogram} \\
(a) & (b) \\
\includegraphics[scale=0.22]{Pictures_B/hist_2_patient_grps} &
\includegraphics[scale=0.22]{Pictures_B/egfr_meseaurements_histogram} \\
(c) & (d)\\
\end{tabular}
\caption{\label{fig:data} Dataset characteristics. (a) Histogram showing the number of eGFR measurements recorded across all patients at each age. (b) Histogram showing the number of eGFR measurements over the range of observed eGFR values. (c) Histogram showing the distribution of the age ranges over which patients had eGFR recordings. (d) Histogram showing the number of eGFR measurements recorded per patient.}
\end{figure}

The dataset used in this work contains the eGFR time series from 488 patients treated at East Kent University Hospital, and was collected as part of a study seeking to understand the characteristics of acute kidney injury and its impact on chronic kidney disease. Each patient’s eGFR time series was labelled as either stable, linear or step-change by five experts. However, for the purpose of this work, we grouped the unstable (linear and step-change) trends together and sought to distinguish only between stable and unstable time series. This is because clinicians are largely concerned with distinguishing between those patients with and without stable eGFR measurements, as this enables them to identify patients who are likely to need further monitoring (those with unstable measurements). Of the 488 patients, 260 (53.3\%) have stable time series and 228 (46.7\%) unstable, while 275 (56.4\%) were male and 213 (43.6\%) female. In total, there were 10,873 eGFR measurements across the 488 patients. Figure~\ref{fig:data} summarises the main characteristics of the dataset. Approximately 95\% of the patients are between the ages of 60 and 90, with eGFR values between 25 and 95 mL/min/ 1.73$m^2$.

\section{Methodology}
\label{method}
\begin{figure*}
\centering
\includegraphics[width=.8\linewidth]{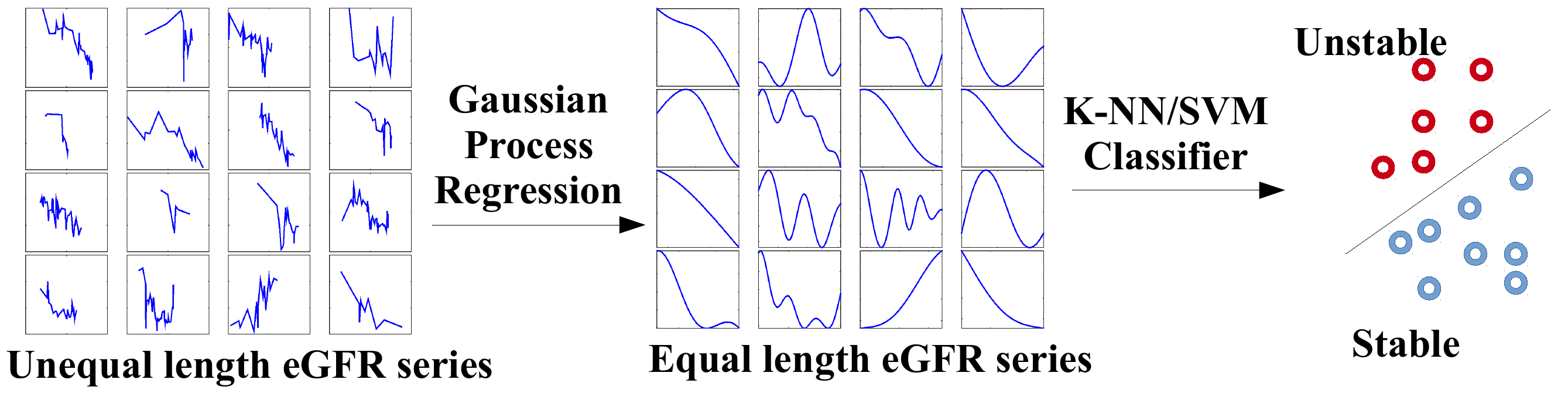}
\caption{Each of the eGFR series of 16 patients (left) is modelled using GPR in order to  produce a fixed-size vector of 50 observations representing the entire observable time series. A collection of these fixed-size vectors (middle) are then given as input to a standard supervised classification algorithm, such as K-NN or SVM, to determine if an eGFR series is stable or unstable (right).}
\label{flowchart}
\end{figure*}

The proposed automatic eGFR-trend classification pipeline consists of a regression step followed by a classification step, as shown in Figure~\ref{flowchart}. The unequal length eGFR series for each patient is modelled using GPR, which produces a fixed-size vector of 50 observations over the patient's recorded age. This fixed-size vector is then given as input to a standard supervised classification algorithm in order to classify the eGFR series as stable or unstable.

\subsection{Gaussian Process Regression}
A Gaussian process (GP) is a collection of random variables, any finite collection of which has a joint Gaussian distribution~\cite{williams2006gaussian}.  GPs characterise the probability  distribution  over  functions  by a specified mean function $\bar{d}(\mb{x})$ and a
covariance function $k(\mb{x},\mb{x}')$~\cite{rasmussen2010gaussian}.


To describe a real process $f(\mb{x})$ as a GP, we write: $f(\mb{x})\sim \mc{GP}\left(\bar{d}(\mb{x}),k(\mb{x},\mb{x}')\right)$. Here $\bar{d}(\mb{x})$  = E$\{d(\mb{x})\}$ and $k(\mb{x},\mb{x}')$ = E$\{(d(\mb{x})-\bar{d}(\mb{x}))(d(\mb{x}')-\bar{d}(\mb{x}'))\}$, where $E\{g(\mb{x})\}$ denotes the expectation of a function $g$ over the variable $\mb{x}$.

Given a set of measurements $\mc{D} = \{(\mb{x}_i,d(\mb{x_i})\}^{N}_{i=1}$, the goal is to estimate the true output $d(x^*)$ at an arbitrary $\mb{x}^*$ given the relation:
\begin{equation}
d^i_{obs} = d(\mb{x_i})+\epsilon(\mb{x_i}); \quad \epsilon(\mb{x_i}) \sim \mc{N}(0,\sigma_n^2)
\end{equation}

The prior distribution of the observed target $d(\mb{x})$ is given by: 
    \begin{equation}
	\label{eq2}
      d(\mb{x}) \sim \mc{N}\left(\bar{d}(\mb{x}),k(\mb{X},\mb{X}'))\right)
    \end{equation}
where $k(\mb{X},\mb{X}')$ is the covariance matrix between all pairs of training points. A squared exponential kernel was used to determine the covariance matrix, where the squared exponential kernel (Gaussian/RBF) is defined as: $\kappa(\mb{x},\mb{x}') =       \exp(\frac{-(\mb{x}-\mb{x}')^2}{2\gamma^2})$, with $\gamma$ the length scale of the kernel.

The distribution of the estimated mean value $\mb{d}({\mb{x}})$ can be computed by conditioning on the training data to get $p(d(\mb{x})|\mb{x}^*,D)$. The joint distribution over $d(\mb{x})$ and the new datapoint $\mb{x}^*$ is computed using:
\scriptsize
    \begin{equation}\nonumber
      \left[
        \begin{array}{c}
          \mb{d}_{obs} \\
          d(\mb{x}^*)
        \end{array}
        \right] 
      \sim \mc{N}
      \left(
      \left[
        \begin{array}{c}
          \bar{d}(\mb{X})\\ 
		  d(\mb{x}^*)
        \end{array}
        \right],
      \left[
        \begin{array}{cc}
          K(\mb{X},\mb{X})+\sigma^2_nI & K(\mb{X},\mb{x}^*) \\
          K(\mb{x}^*,\mb{X}) & k(\mb{x}^*,\mb{x}^*)
        \end{array}
        \right]
      \right)
    \end{equation}
\normalsize
Here,  $\mb{d}_{obs} = \left(d^1_{obs},\hdots,d^N_{obs}\right)^T$; $\mb{X} = \{\mb{x}_1,\hdots,\mb{x}_N\}$, $\bar{\mb{d}}(\mb{X})_i = \bar{d}(\mb{x}_i)$, and $K(\mb{X},\mb{X})_{ij} = k(\mb{x}_i,\mb{x}_j)$.

The conditional distribution of Equation~\ref{eq2} allows us to get the distribution of 
$d(\mb{x}^*)$ with the following mean and covariance:
\begin{equation}
    d(\mb{x}^*) \sim \mc{N}\left(E\{d(\mb{x}^*)\},var\{d(\mb{x}^*)\}\right)
    \end{equation}
    where
\scriptsize
    \begin{eqnarray}\nonumber
      E\{d(\mb{x}^*)\} &=& \underbrace{\bar{d}(\mb{x}^*)}_{\text{prior}} + K(\mb{x}^*,\mb{X})\left[K(\mb{X},\mb{X})+\sigma^2_nI\right]^{-1}(\mb{d}_{obs}-\bar{\mb{d}}(\mb{X})) \\
        var\{d(\mb{x}^*)\} &=& \underbrace{k(\mb{x}^*,\mb{x}^*)}_{\text{prior}} - K(\mb{x}^*,\mb{X})\left[K(\mb{X},\mb{X})+\sigma^2_nI\right]^{-1}K(\mb{X},\mb{x}^*) \nonumber
    \end{eqnarray}
\normalsize
The GPstuff toolbox~\footnote{\url{http://research.cs.aalto.fi/pml/software/gpstuff/}}~\cite{vanhatalo2013gpstuff}, was used for modelling the eGFR time series, with the hyperparameters for the squared exponential kernel tuned using maximum a posteriori estimates.

\subsection{Classification Details \& Performance Evaluation}

The fixed size eGFR series can be used as the features for any standard classification algorithm. Here we have used K-NNs and SVMs for this purpose, as they are  known  for  their high  classification  accuracy. The Euclidean distance measure was used for the K-NN classifier, with K = 3. For  the SVMs, the radial basis function kernel was used with  $\sigma =10$. Performance was evaluated using 5-fold cross validation, with the performance for each fold evaluated using the F-score. 

Let $\mathcal{Y}$ be the classifier's prediction and $\omega$ the groundtruth. Then the stable/unstable classification of an eGFR series is made on the following basis:
 \begin{equation}
   \label{equ:decision_func}
   \mbox{decision}(\mathcal{Y}) = \left\{ \begin{array}
       {r@{\quad \quad}l}
       stable & \mbox{if }  y =1  \\
       unstable & \mbox{otherwise},
     \end{array} \right.
 \end{equation}

This can result in the following outcomes: (i) a true positive ($\mbox{TP} \equiv y = 1, \omega = 1;$), (ii) a false positive ($\mbox{FP} \equiv y = 1, \omega = 0;$), (iii) a true negative ($\mbox{TN} \equiv y = 0, \omega = 0;$) and (iv) a false negative ($\mbox{FN} \equiv y = 0, \omega = 1;$). Using the rates of these four outcomes, recall and precision can be derived. Recall describes the completeness of the classification, and precision the actual accuracy of the classification. They are defined with recall =$\frac{TP}{TP+FN}$ and precision =$\frac{TP}{TP+FP}$. While recall and precision can be individually used to determine the quality of a classifier, it is often more convenient to have a single measure. The F-score achieves this by combining the recall and precision in a single equation:

\[ F = 2*\frac{precision*recall}{precision+recall} \] 


\subsection{Groundtruth}

For each eGFR time-series, five nephrologists annotated with one of three labels: stable, linear or a step-change. Here linear and step-change are both considered to be unstable trends. The class label assigned to an eGFR time-series is based on the consensus of these five annotations. In order to evaluate the performance of an individual expert, we compared their labels to those of the remaining experts, and from this derived the F-score for the given expert. The performance of the five experts can be seen in Table~\ref{exp_res}.

\begin{table}[tb]
\centering
\caption{The F-score of each expert of the five experts and their mean}
\begin{tabular}{|l|l|l|l|l|l|l|}
\hline
\textbf{}          & \textbf{E1} & \textbf{E2} & \textbf{E3} & \textbf{E4} & \textbf{E5} & \textbf{Mean} \\ \hline\textbf{F-score} & 0.8417            & 1                 & 1                 & 1                 & 0.9938            & \textbf{0.9671}  \\ \hline
\end{tabular}
\label{exp_res}
\end{table}

\section{Experiments \& Results}
\label{exp}
Our experimental procedures and objectives were as follows:
\begin{itemize}
\item {\bf Classification using derived statistics:} In order to provide a baseline for evaluating the performance benefits of equalising the lengths of the eGFR time series, we investigated whether it was possible to classify eGFR trends using only basic statistics derived from the time series.
\item {\bf Classification using equalised eGFR time series:} The effects of equalising the lengths of the eGFR time series were evaluated using two approaches. First, we equalised each patient's time series while taking into account age, by fitting a GPR model and uniformly resampling 50 datapoints from the expected trend between the ages of 30 and 90. Next we used the fitted GPR model to uniformly resample 50 datapoints for each patient, with the sampling restricted to the age range in which the patient had eGFR measurements recorded. Although this ensures that the new datapoints are only generated from the age range with low variance, it may potentially cause the sampling rate to be different for each patient.
\item {\bf Classification using derived statistics and equalised eGFR time series:} Classification using only datapoints resampled from GPR can be considered to be an instance of pattern matching, as trends are classified purely based on similarities between GPR models. Therefore, we investigate whether reintroducing information from derived statistics, such as the age range over which measurements were taken, is beneficial.
\item {\bf Comparison with linear interpolation:} Finally, we compare the GPR-based classification method with one where the 50 datapoints are produced by linear interpolation. This experiment examines whether the smoothness of the fitted curve provided by GPR warrants performance generalisation or not.
\end{itemize}

\subsection{Derived Statistics}
\begin{figure}[ht]
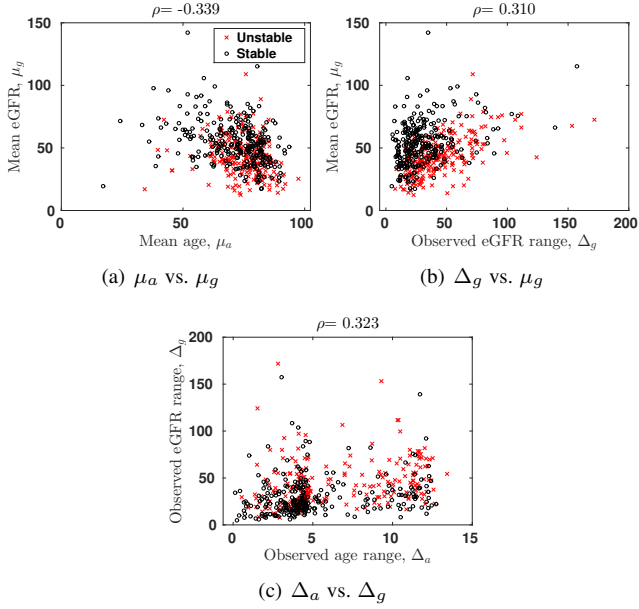

\centering
\subfigure[$\mu_a$ vs. $\mu_g$ ]
{\includegraphics[scale=0.22]{Pictures_B/stats_corr_age_mean_vs_eGFR_mean.eps}}
\subfigure[$\Delta_g$ vs. $\mu_g$ ]
{\includegraphics[scale=0.22]{Pictures_B/stats_corr_eGFR_range_vs_eGFR_mean.eps}}
\subfigure[$\Delta_a$ vs. $\Delta_g$ ]
{\includegraphics[scale=0.22]{Pictures_B/stats_corr_age_range_vs_eGFR_range.eps}}
\caption{\label{fig:statistics} A scatter plot of different derived statistics. }
\label{derived_stats}
\end{figure}

For each patient, we derived four statistics from their series of eGFR measurements: the age range over which eGFR measurements were made ($\Delta_a$), the range of the eGFR measurement values ($\Delta_g$), the mean age at which a patient had their eGFR measured ($\mu_a$) and their mean eGFR value ($\mu_g$). From Figure~\ref{fig:statistics}(a) we can see that $\mu_a$ is negatively correlated with $\mu_g$, showing that as patients get older their eGFR tends to decline. In addition, patients with an unstable trend tend to have a lower $\mu_g$ value. Patients with a greater $\mu_g$ also tend to have their eGFR measured over a longer age range (Figure~\ref{fig:statistics}(b)), possibly because a greater mean eGFR value means that the patient is less likely to have suffered from complications that cause their eGFR to stop being measured. As patients with unstable eGFR trends tend to have both a greater $\mu_a$ and lower $\mu_g$ value than patients with stable trends, the derived statistics are likely to provide some discriminative power to distinguish between the two trends. In order to ascertain the level of this discriminative power, we fed the statistics to K-NN and SVM for classification. 
We found that this baseline approach gives an average F-score of 0.71 based on 5-fold cross validation (Table~\ref{results}).

\subsection{eGFR Time Series Equalisation}
\begin{figure}[ht]
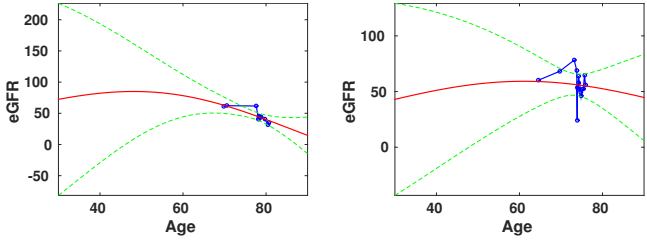

\centering
\begin{tabular}{cc}
\includegraphics[scale=0.22]{Pictures_B/GPR_30-905.eps} & \includegraphics[scale=0.22]{Pictures_B/GPR_30-902.eps}\\
\end{tabular}
\caption{eGFR resampled between ages 30 and 90 for unstable and stable trends. An increase in variance is seen when eGFR readings are not present.}
\label{30-90}
\end{figure}

Using the GPR model learned from each patient's eGFR time series, we resampled 50 uniformly spaced datapoints between the ages of 30 and 90 for each patient. This ensures that not only are all patient's resampled time series of equal length, but also that the interval between measurements is the same. However, while the model shows relatively low variance in the age range where a patient has measurements, the variance increases markedly outside this range. For example, the patient in Figure~\ref{30-90} had their eGFR recorded between the ages of 55 and 75, and consequently the GPR model fit using their measurements shows lower variance  within this age range. The performance of the K-NN classifier and SVM trained using the resampled datapoints was 0.57 and 0.43 respectively; lower than that achieved using only derived statistics (Table~\ref{results}).

\begin{figure}[ht]
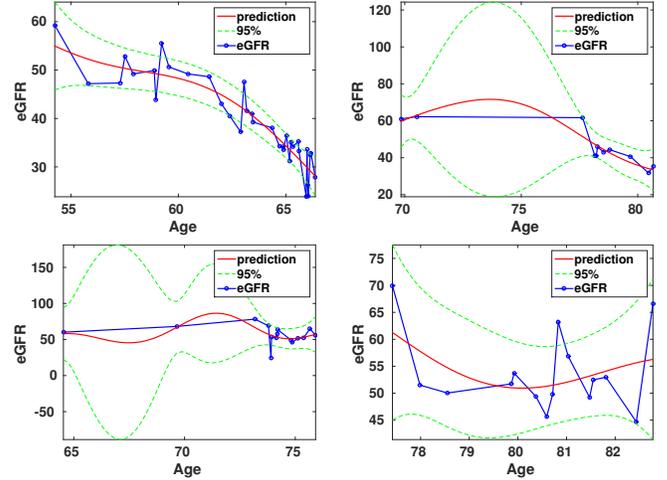

\centering
\begin{tabular}{cc}
\includegraphics[scale=0.22]{Pictures_B/GPR_1.eps} & \includegraphics[scale=0.22]{Pictures_B/GPR_5.eps}\\
\includegraphics[scale=0.22]{Pictures_B/GPR_2.eps} & \includegraphics[scale=0.22]{Pictures_B/GPR_4.eps}\\
\end{tabular}
\caption{Unequal length eGFR series (blue) modelled through GPR to produce a fixed-size vector of 50 observations (red) over the patients recorded age. (Top) figure shows the unstable trend and (bottom) figure shows the stable trend.}
\label{egFRGPR}
\end{figure}

For each patient, we also used the GPR model learned from their eGFR measurements to resample 50 uniformly spaced datapoints within the age range over which they have eGFR measurements recorded. The fitted mean curve, along with the 95\% variance interval, for four patients can be seen in Figure~\ref{egFRGPR}. When compared to the those resampled over the entire 30 to 90 year age range, the resampled curves within the age range can be expected to have lower variance. This decrease in variance is likely responsible for the substantial improvement in performance achieved by both the K-NN classifier and SVM when compared to their performance using data resampled between ages 30 and 90 (Table~\ref{results}).

\subsection{Combining Statistics \& Equalisation}
In the previous GPR experiments, only the resampled eGFR values are taken into account, thereby equating the problem of trend classification to that of pattern matching, since the notion of time is not considered. As we believe that age information is likely to be important in classifying the trends, we reintroduce this information by creating a feature vector for each patient that consists of the four derived statistics and the 50 uniformly resampled datapoints within the age range over which the patient has eGFR measurements. By reintroducing the derived statistics, the F-score increased from 0.87 to 0.90 for the K-NN classifier, and from 0.86 to 0.89 for the SVM (Table~\ref{results}).

\subsection{Comparison with Interpolation} 
\begin{figure}[ht]
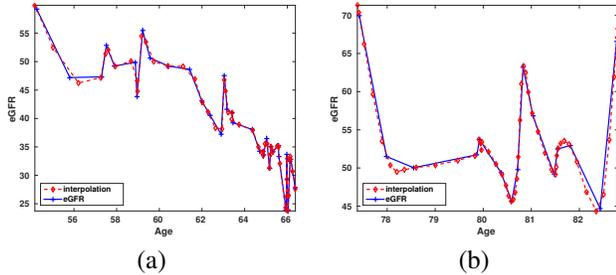

\centering
\begin{tabular}{cc}
\includegraphics[scale=0.22]{Pictures_B/Bint1.eps} & \includegraphics[scale=0.22]{Pictures_B/Bint4.eps}\\
(a) & (b)
\end{tabular}
\caption{Unequal length eGFR series (blue) modelled using linear interpolation to produce a fixed-size vector of 50 observations (red) over the range for which a patient has eGFR measurements. The figure shows both (a) unstable  and (b) stable trends.}
\label{bint}
\end{figure}

Rather than using GPR for equalising the eGFR lengths, we could have used linear interpolation. This allows the structure and trend of the eGFR series to be preserved, while still producing fixed size vectors (as seen in Figure~\ref{bint}). A K-NN classifier and SVM trained on data generated in this manner had an F-score of 0.84 and 0.87 respectively. When coupled with the derived statistics, the F-score of the K-NN classifier improved to 0.86, while the F-score of the SVM remained the same. However, results for both classifiers were lower than for equalisation performed using GPR, possibly due to the smoother fitted curve produced by GPR.

\begin{table*}
{\small
\centering
\caption{Classifier F-score}
\label{results}
\begin{tabular}{c|c|c|c|c|c|c|c|c|c|c|c|p{1cm}|}
\cline{2-13}
                                                & \multicolumn{2}{p{2cm}|}{\textbf{30-90 GPR}} & \multicolumn{2}{p{2cm}|}{\textbf{Statistics}} & \multicolumn{2}{c|}{\textbf{GPR}} & \multicolumn{2}{p{2cm}|}{\textbf{Statistics + GPR}}      & \multicolumn{2}{c|}{\textbf{Interpolation}} & \multicolumn{2}{p{2cm}|}{\textbf{Statistics + Interpolation}} \\ \cline{2-13} 
                                                & \textbf{K-NN}      & \textbf{SVM}       & \textbf{K-NN}       & \textbf{SVM}       & \textbf{K-NN}   & \textbf{SVM}    & \textbf{K-NN}            & \textbf{SVM}             & \textbf{K-NN}        & \textbf{SVM}         & \textbf{K-NN}               & \textbf{SVM}               \\ \hline
\multicolumn{1}{|c|}{\textbf{Fold-1}}           & 0.5952             & 0.4450              & 0.7526              & 0.7465             & 0.8950           & 0.8643          & 0.9258                   & 0.8443                   & 0.8762               & 0.8858               & 0.8762                      & 0.8760                      \\ \hline
\multicolumn{1}{|c|}{\textbf{Fold-2}}           & 0.5918             & 0.4368             & 0.6927              & 0.6632             & 0.8451          & 0.8450           & 0.9446                   & 0.9151                   & 0.7343               & 0.8162               & 0.8670                       & 0.8776                     \\ \hline
\multicolumn{1}{|c|}{\textbf{Fold-3}}           & 0.5143             & 0.4313             & 0.7427              & 0.8151             & 0.9077          & 0.8732          & 0.9178                   & 0.9065                   & 0.8976               & 0.9183               & 0.8776                      & 0.8876                     \\ \hline
\multicolumn{1}{|c|}{\textbf{Fold-4}}           & 0.5970              & 0.4121             & 0.7142              & 0.7245             & 0.8850           & 0.8757          & 0.8542                   & 0.8958                   & 0.8164               & 0.8234               & 0.8024                      & 0.8182                     \\ \hline
\multicolumn{1}{|c|}{\textbf{Fold-5}}           & 0.5633             & 0.4343             & 0.6561              & 0.6390              & 0.8955          & 0.8655          & 0.8645                   & 0.8658                   & 0.8792               & 0.9046               & 0.8622                      & 0.8752                     \\ \hline
\multicolumn{1}{|c|}{\textit{\textbf{Average}}} & \textit{0.5723}    & \textit{0.4319}    & \textit{0.7116}     & \textit{0.7176}    & \textit{0.8856} & \textit{0.8647} & \textit{\textbf{0.9013}} & \textit{\textbf{0.8855}} & \textit{0.8407}      & \textit{0.8696}      & \textit{0.8570}             & \textit{0.8669}            \\ \hline
\end{tabular}
}
\end{table*}

\section{Conclusion \& Discussion}
\label{conclusion}
Due to the significance of kidney function in many chronic conditions, it is important for clinicians to be able to quickly and accurately screen patients' eGFR time series to identify those whose kidney function is at risk of deteriorating. Given that many patients with long term conditions are managed in their general practices, the screening process would ideally be automated and able to remotely monitor patients. Despite this, no automated methods exist that solve this problem. Our work is therefore the first reported attempt to design an automated process and compare it with human experts.

In spite of the complexity of eGFR trends, we found that it was possible to utilise a supervised machine learning approach to automatically determine if an eGFR trend is stable or unstable. We found that by equalising the lengths of the eGFR time series for each patient by uniformly resampling new datapoints, our approach could classify the eGFR trend with an F-score approaching that of human experts (0.90 compared to 0.97 for the experts). The best approach for performing this equalisation is to fit a GPR model, and then resample new datapoints from within the age range of the patient where eGFR measurements are observable. This approach performs better than resampling new datapoints from the fitted GPR across the same age range (30 to 90 years) for all patients, and better than using linear interpolation to resample the new datapoints. This shows that smoothness of the fitted curve plays an important in performance generalisation.
Finally, the inclusion of statistics derived from the original eGFR time series of a patient further improves the classification performance for all classifiers except the SVM trained on interpolated data.




%
{\small
\bibliographystyle{IEEEbib}
\bibliography{santosh_references,eGFRclass}
}

\end{document}